\documentclass{article}
\usepackage{ijcai24}

\usepackage{times}
\usepackage{soul}
\usepackage{url}
\usepackage[hidelinks]{hyperref}
\usepackage[utf8]{inputenc}
\usepackage[small]{caption}
\usepackage{graphicx}
\usepackage{amsmath}
\usepackage{amsthm}
\usepackage{booktabs}
\usepackage{algorithm}
\usepackage{algorithmic}
\usepackage[switch]{lineno}
\usepackage{multirow}
\usepackage{siunitx}

\title{VN-Net: Vision-Numerical Fusion Graph Convolutional Network for Sparse \\
Spatio-Temporal Meteorological Forecasting}

\author{
Yutong Xiong$^{1,\dag}$\and
Xun Zhu$^{1,\dag}$\and
Ming Wu$^{1,\ddag}$\and
Weiqing Li$^1$\and
Fanbin Mo$^1$\and\\
Chuang Zhang$^{1}$\And
Bin Zhang$^1$
\affiliations
$^1$Beijing University of Posts and Telecommunications, China \\
$^\dag$Equal contribution \quad $^\ddag$Corresponding author
\emails
\{2017210039, zhuxun, wuming, bupt\_weiqing, noah227, zhangchuang, bluezb\}@bupt.edu.cn
}

\begin{document}

\maketitle

\begin{abstract}
    Sparse meteorological forecasting is indispensable for fine-grained weather forecasting and deserves extensive attention.
    Recent studies have highlighted the potential of spatio-temporal graph convolutional networks (ST-GCNs) in predicting numerical data from ground weather stations. 
    However, as one of the highest fidelity and lowest latency data, the application of the vision data from satellites in ST-GCNs remains unexplored.
    There are few studies to demonstrate the effectiveness of combining the above multi-modal data for sparse meteorological forecasting.
    Towards this objective, we introduce Vision-Numerical Fusion Graph Convolutional Network (VN-Net), which mainly utilizes: 
    1) Numerical-GCN (N-GCN) to adaptively model the static and dynamic patterns of spatio-temporal numerical data; 
    2) Vision-LSTM Network (V-LSTM) to capture multi‐scale joint channel and spatial features from time series satellite images; 
    3) Double Query Attention Module (DQAM) to conduct multi-modal feature interaction; 
    4) a GCN-based decoder to generate hourly predictions of specified meteorological factors. 
    As far as we know, VN-Net is the first attempt to introduce GCN method to utilize multi-modal data for better handling sparse spatio-temporal meteorological forecasting.
    Our experiments on Weather2K dataset show VN-Net outperforms state-of-the-art by a significant margin on mean absolute error (MAE) and root mean square error (RMSE) for temperature, relative humidity, and visibility forecasting.
    Furthermore, we conduct interpretation analysis and design quantitative evaluation metrics to assess the impact of incorporating vision data.

\end{abstract}

\section{Introduction}
Weather forecasting is a long-standing scientific challenge with practical significance, which is closely relevant to human society and economic activities.
Physics-based Numerical Weather Prediction (NWP) models have driven the main forecasts that are available worldwide in the past few decades.
NWP models collect and process diverse dense and sparse meteorological data sources through sophisticated data assimilation, forming the initial dense atmospheric representation, which is then approximated and extended to the future using mathematical and physical equations~\cite{bougeault2010thorpex}.
But limitations arise from the existing atmospheric physics theories and computational costs~\cite{bauer2015quiet}.

Recently, data-driven approaches that directly solve forecasting tasks by learning a functional mapping using deep neural networks have garnered considerable attention and achieved prominent breakthroughs in medium-range global weather forecasting~\cite{bi2023accurate,lam2023learning,Chen2023FuXiAC,chen2023fengwu}.
However, these methods rely on curated reanalysis dataset ERA5~\cite{hersbach2020era5} during training, which are mainly limited to regular-grid forecasting.
In addition, the construction of the reanalysis data depends on data assimilation of NWP~\cite{mooney2011comparison}, which is a black box.
There is also a significant mismatch between ground truth observations and the values of the same variables provided by the NWP states~\cite{andrychowicz2023deep}.
Different from global weather forecasting, sparse meteorological forecasting is crucial for fine-grained weather forecasting and expected to learn from sparse observation data, ranging from time series individual target to spatio-temporal unstructured targets.
Compared with reanalysis data, sparse observation data explicitly benefits from high fidelity, low latency, and low collection cost.
Although latest studies~\cite{ma2023histgnn,zhu2023weather2k} have highlighted the potential of spatio-temporal graph convolutional networks (ST-GCNs) in predicting numerical data from ground weather stations, multi-modal sparse meteorological forecasting is still a thorny challenge due to significant modal differences, sparsity, and heterogeneity among observation data sources.

\nocite{bougeault2010thorpex}
\nocite{bauer2015quiet}
\nocite{bi2023accurate}
\nocite{lam2023learning}
\nocite{Chen2023FuXiAC}
\nocite{chen2023fengwu}
\nocite{hersbach2020era5}
\nocite{mooney2011comparison}
\nocite{andrychowicz2023deep}

In this work, we focus on sparse spatio-temporal meteorological forecasting using multi-modal data collected from ground weather stations and meteorological satellites, both of which are considered as the highest fidelity and lowest latency data.
Towards this objective, we propose Vision-Numerical Fusion Graph Convolutional Network (VN-Net).
VN-Net enhances the performance of sparse meteorological forecasting from the perspectives of numerical and vision feature extraction and multi-modal fusion.
For numerical data, we introduce Numerical Graph Convolutional Network (N-GCN), which adaptively learn correlations among stations, considering both temporal factors and dynamic relationships in the input.
For vision data, we employ Vision-LSTM Network (V-LSTM) with Multi-Scale joint Channel and Spatial Module (MSCSM) to effectively extract multi-channel information and capture meteorological activities at different scales from time series satellite images.
We further propose Double Query Attention Module (DQAM) to integrate features from numerical and vision branches and then feed the fusion features into a GCN-based decoder to generate the final predictions.
In summary, the main contributions of this paper can be outlined as follows:
\begin{itemize}
\item We are the first to utilize multi-modal observation data collected from ground weather stations and meteorological satellites based on GCN method to handle sparse spatio-temporal meteorological forecasting.
\item VN-Net achieves state-of-the-art on Weather2K dataset, which demonstrates the effectiveness of VN-Net and that incorporating vision data with numerical data can significantly improve forecasting performance.
\item For model interpretation, we conduct attribution analysis of multivariate meteorological factors and design the metrics of Meteorological Factors Contribution (MFC) and Static Information Contribution (SIC) to assess the difference between multi-modal and uni-modal inputs.
\end{itemize}

\section{Related Work}
\subsection{Sparse Meteorological Forecasting}
Sparse meteorological forecasting includes time series forecasting for individual target and spatio-temporal forecasting for unstructured targets.
On the one hand, numerous efforts have been dedicated to solving time series meteorological forecasting by using transformer variants~\cite{zhou2021informer,wu2021autoformer,zhou2022fedformer,yu2023dsformer}, linear models~\cite{zeng2023transformers}, graph evolution learning~\cite{spadon2021pay}, and Large Language Models~\cite{zhou2023one}.
On the other hand, the excellent ability of GNN to model unstructured data is well demonstrated in the forecasting task of PM2.5~\cite{su2023effective}, air quality~\cite{han2022semi}, heatwave~\cite{li2023regional}, rainfall~\cite{zhang2023st}, sea temperature~\cite{kim2023spatiotemporal} and so on.
CLCRN ~\cite{lin2022conditional} proposed graph-based conditional convolution for imitating the meteorological flows to construct the local spatial patterns.
HiSTGNN~\cite{ma2023histgnn} incorporated an adaptive hierarchical graph learning module to capture hidden correlations among meteorological factors in diﬀerent regions.
MFMGCN~\cite{zhu2023weather2k} focused on multi-graph construction and fusion strategy, combining static and dynamic graphs to capture the intrinsic correlation among geographic locations based on historical and actual input data.
Notably, N-GCN in VN-Net advances in spatio-temporal dependency modeling through combining static and dynamic graph learning and incorporating time factors into embedding learning.

\nocite{zhou2021informer}
\nocite{zhou2022fedformer}
\nocite{wu2021autoformer}
\nocite{zhou2023one}
\nocite{chang2023llm4ts}
\nocite{yu2023dsformer}
\nocite{zeng2023transformers}
\nocite{spadon2021pay}
\nocite{su2023effective}
\nocite{ma2023histgnn}
\nocite{li2023regional}
\nocite{zhang2023st}
\nocite{han2022semi}
\nocite{kim2023spatiotemporal}
\nocite{zhu2023weather2k}
\nocite{lin2022conditional}

\subsection{Multi-modal Meteorological Forecasting}
The development of models capable of learning from diverse data modalities can effectively enhance the forecasting performance~\cite{chen2023foundation}.
In the field of solar irradiance forecasting, recent studies explored the combination of satellite cloud images~\cite{shan2023deep,shan2022ensemble}, infrared cloud images~\cite{ajith2021deep}, sky images~\cite{zhang2023advanced,liu2023transformer} with historical ground monitoring data from distributed stations.
BFM~\cite{yu2021application} used a fusion machine learning model based on four types of meteorological data for visibility forecasting.
For tropical cyclone forecasting, Hurricast~\cite{boussioux2022hurricane} employed XGBoost to make predictions using statistical features based on historical data and spatial–temporal features extracted with encoder–decoder architectures from atmospheric reanalysis maps.
MGTCF~\cite{huang2023mgtcf} proposed a GAN-based model, which consists of 2D and 1D data encoders to process tropical cyclone attributes and meteorological grid data, Env-Net to extract the environment information, and GC-Net to choose accurate tendency predicted by multiple generators.
MMST-ViT~\cite{lin2023mmst} designed spatial, temporal and multi-modal transformer modules to process satellite images and meteorological data to capture the impacts of short-term weather variations and long-term climate change on crop yield prediction.
MTTF~\cite{cao2023mttf} proposed a multi-modal transformer method that incorporates time series observation data and ERA5 temperature data for temperature forecasting.
MMSTP~\cite{zhang2023mmstp} designed an encoder module that combines the advantages of CNN and transformer and cross temporal self-attention mechanism to process satellite images and radar echo maps for precipitation forecasting.
Differently, our VN-Net is the first attempt to provide a multi-modal GCN method for spatio-temporal meteorological forecasting.

\nocite{chen2023foundation}
\nocite{shan2023deep}
\nocite{zhang2023advanced}
\nocite{lin2023mmst}
\nocite{liu2023transformer}
\nocite{cao2023mttf}
\nocite{huang2023mgtcf}
\nocite{zhang2023mmstp}
\nocite{boussioux2022hurricane}
\nocite{ajith2021deep}
\nocite{yu2021application}

\begin{figure*}
\centerline{\includegraphics[scale=0.305]{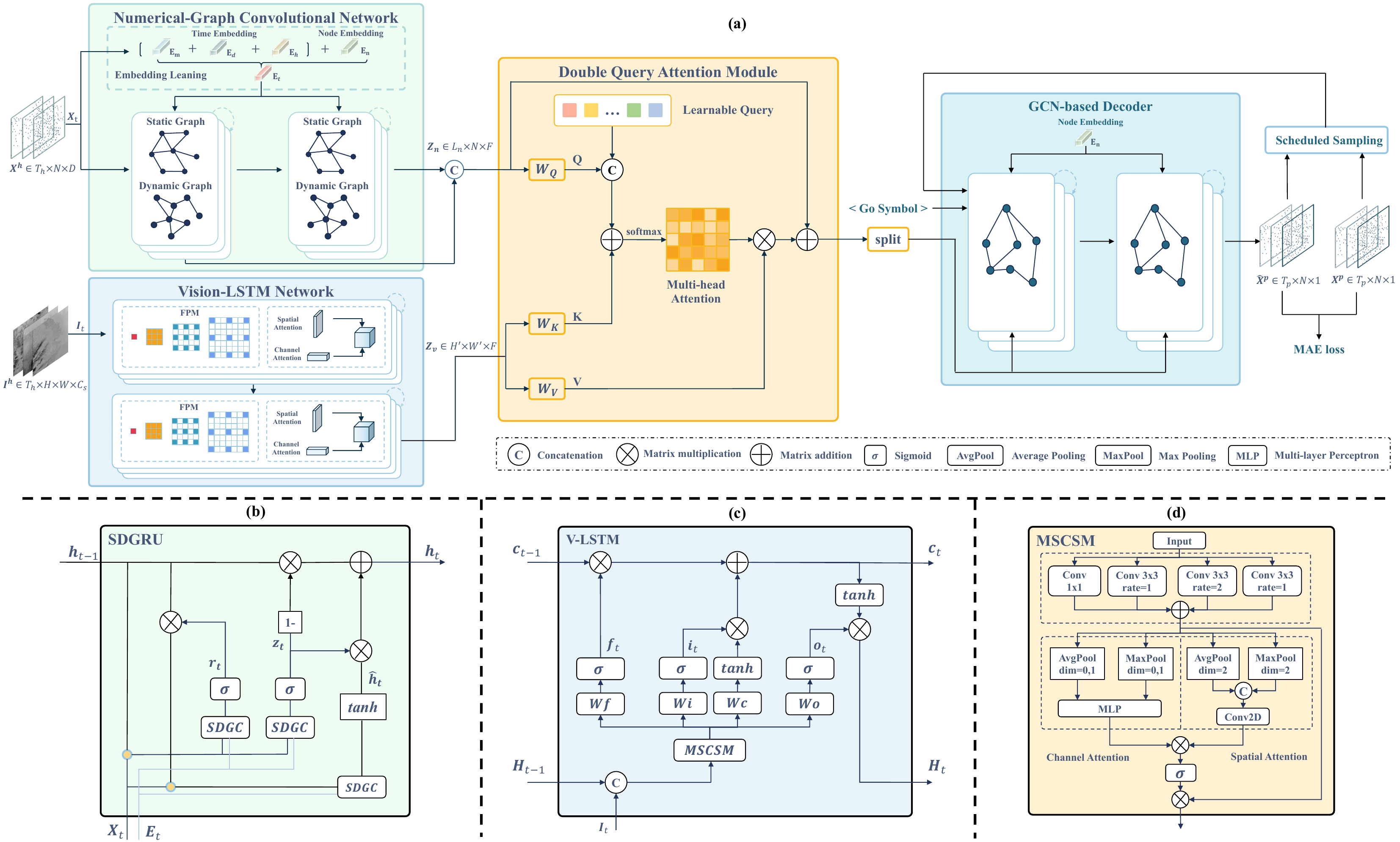}}
\caption{The architecture and key modules of VN-Net. 
(a) Overall design.
(b) SDGRU.
(c) V-LSTM.
(d) MSCSM.}
\label{F1}
\end{figure*}

\section{Method}
In this section, we introduce the details of VN-Net, which consists of Numerical Graph Convolutional Network (N-GCN), Vision-LSTM Network (V-LSTM), Double Query Attention Module (DQAM), and GCN-based Decoder, as illustrated in Figure~\ref{F1} (a).

\subsection{Task Setting}
We first formalize the problem setting of multi-modal sparse spatio-temporal meteorological forecasting.
The meteorological network in a certain region can be represented as graph 
$G =\left(V\text{, } E\text{, } \mathbf{A}  \right) $, where $V$ is a set of $N$ nodes representing ground weather stations in different geographic locations, $E$ is a set of edges, 
and $\mathbf{A} \in \mathbf{R}^{N\times N}$ is a weighted adjacency matrix representing the nodes proximity.
Let $\mathbf{X}_t\in \mathbf{R}^{N\times D}$ and $\mathbf{I}_t \in \mathbf{R}^{H\times W\times C_s}$ represent the observed numerical meteorological factors of all stations and the vision image data of satellite in corresponding region at time step $t$, 
where $D$, $C_s$, $H$ and $W$ are the number of meteorological factor, channel, height and width of the satellite image, respectively. 
$\mathbf{X}_t^{f*} \in \mathbf{R}^{N \times 1}$ is the values of meteorological factor $f*$ of all nodes.
Given $T_h$ steps historical data $\mathbf{X}^h=[\mathbf{X}_{t-{T_h}+1},\cdots,\mathbf{X}_t]$ and $\mathbf{I}^h=[\mathbf{I}_{t-{T_h}+1},\cdots,\mathbf{I}_t]$, the task of sparse meteorological forecasting aims to predict the future $T_p$ signals $\mathbf{X}^p_{(f*)}=[\mathbf{X}_{t+1}^{f^*},\cdots,\mathbf{X}_{t+T_p}^{f*}]$ by function $\theta$:
\begin{equation}
\left[\mathbf{X}^h,\mathbf{I}^h;G\right] \stackrel{\theta}{\longrightarrow}[\mathbf{X}^p_{(f*)};G]
\end{equation}

\subsection{Numerical Graph Convolutional Network}
\subsubsection{Spatial Dependency Modeling}
Numerical encoder adopts the general graph convolution framework based on the Graph Laplacian proposed by ~\cite{bruna2013spectral} and using Chebyshev polynomial approximation to realize eigenvalue decomposition~\cite{kipf2016semi}. In practice, N-GCN employ Static-Dynamic Graph Convolution (SDGC), consisting of two parts: a graph convolution approximated by $1^{st}$ order Chebyshev polynomial and a dynamic information propagation determined by input signals:
\begin{equation}
\mathbf{Z}=\left(\mathbf{I}_N+\mathbf{D}^{-\frac{1}{2}}\mathbf{A}\mathbf{D}^{-\frac{1}{2}}\right)\mathbf{X}\mathbf\Theta_1+\mathbf{R}_d\mathbf{X}\mathbf\Theta_2+\mathbf{b}
\end{equation}
where $\mathbf{A}\in \mathbf{R}^{N\times N}$ is the adjacent matrix of the graph, $\mathbf{D}$ is the degree matrix of $\mathbf{A}$, $\mathbf{R}_d\in \mathbf{R}^{N\times N}$ is a matrix capturing dynamic correlation,
$\mathbf{X}\in \mathbf{R}^{N\times C}$ and $\mathbf{Z}\in \mathbf{R}^{N\times F}$ are input and output of the GCN layer, $\mathbf\Theta_1,\mathbf\Theta_2\in \mathbf{R}^{C\times F}$ and $\mathbf{b}\in \mathbf{R}^F$ represent the learnable weights and bias.

\subsubsection{Embedding Learning}
In embedding learning, we first randomly initialize a learnable node embedding matrix $\mathbf{E}_{n}\in \mathbf{R}^{N\times d_e}$, where each row of $\mathbf{E}_{n}$ represents the embedding of a node and $d_e$ denotes the dimension of the node embedding.
The time embedding $\mathbf{E}_{m},\mathbf{E}_{d},\mathbf{E}_{h}\in \mathbf{R}^{{T_h}\times d_e}$ are determined by the time step $[t-{T_h}+1,\cdots,t]$ of input data, representing time information of month, day, and hour, respectively.
The final embedding is formulated as:
\begin{equation}
\mathbf{E}=\mathbf{E}_{n}+\mathbf{E}_{m}+\mathbf{E}_{d}+\mathbf{E}_{h}
\end{equation}
where broadcasting mechanism has been employed thus $\mathbf{E}\in \mathbf{R}^{{T_h}\times N\times d_e}$.
Note that the embedding $\mathbf{E}_t\in \mathbf{R}^{N\times d_e}$ at time step $t$ is different due to the time embedding.

\subsubsection{Static Graph Learning}
In each time step $t$, the static spatial dependencies between each pair of nodes is calculated by the embedding $\mathbf{E}_t$ :
\begin{equation}
\mathbf{D}^{-\frac{1}{2}}\mathbf{A}\mathbf{D}^{-\frac{1}{2}}=softmax(ReLU(\mathbf{E}_t\cdot\mathbf{E}_t^T))
\end{equation}

\subsubsection{Dynamic Graph Learning}
Considering the impact of rapid weather changes on inter-node relationships, the dynamic spatial dependencies can be inferred by input signal $\mathbf{X}_t$:
\begin{equation}
\mathbf{R}_d=softmax(ReLU((\mathbf{X}_t\mathbf{W}_1)\cdot(\mathbf{X}_t\mathbf{W}_2)^T))
\end{equation}
where $\mathbf{W}_1$ and $\mathbf{W}_2$ are parameters of linear layers.

\subsubsection{Node Adaptive Parameter Learning}
Spatio-temporal meteorological forecasting is a complex issue, with variations in the utilization of different features at each geographical location.
To this end, the Node Adaptive Parameter Learning (NAPL) module is employed to maintain a unique parameter space for each node to learn node-specific patterns.
Taking $\mathbf\Theta_1$ as an example for illustration, assigning parameters for each node results in $\mathbf\Theta_1\in \mathbf{R}^{N\times C\times F}$, which is too huge to optimize and would lead to overfitting.
Based on matrix decomposition, NAPL learns the weight matrix $\mathbf\Theta_1$ through node embedding $\mathbf{E}_{n}$ and a weight pool $\mathbf{W}_s\in R^{d_e\times C\times F}$, i.e. $\mathbf\Theta_1=\mathbf{E}_n\cdot\mathbf{W}_s$.
Note that the number of parameters have a significant reduction because $d_e\ll N$.
The same operation is implemented for $\mathbf\Theta_2$ and $\mathbf{b}$.
The final GCN operation can be formulated as:
\begin{equation}
\resizebox{.91\linewidth}{!}{$
\mathbf{Z}=\left(\mathbf{I}_N+\mathbf{D}^{-\frac{1}{2}}\mathbf{A}\mathbf{D}^{-\frac{1}{2}}\right)
\mathbf{X}\mathbf{E}_n\mathbf{W}_s +\mathbf{R}_d\mathbf{X}\mathbf{E}_n\mathbf{W}_d  +\mathbf{E}_n\mathbf{W}_b
$}
\end{equation}

\subsubsection{Temporal Dependency Modeling}
We utilize Gated Recurrent Units (GRU) proposed by ~\cite{chung2014empirical} to capture temporal dependency.
As shown in Figure~\ref{F1} (b), we introduce a Static-Dynamic Graph Convolution Gated Recurrent Unit (SDGRU) by replacing the MLP layers in the GRU with our static and dynamic graph convolution and NAPL methods.
The calculation process in SDGRU is as follows:
\begin{equation}
\begin{aligned}
\tilde{\mathbf{A}}_s&=\mathbf{I}_N+softmax(ReLU(\mathbf{E}_t\cdot\mathbf{E}_t^T)  \\
\tilde{\mathbf{A}}_d&=softmax(ReLU(\mathbf{M}_1\cdot\mathbf{M}_2^T)) \\
\mathbf{z}_t \,&=\sigma (\tilde{\mathbf{A}}_s[\mathbf{X}_t,\mathbf{h}_{t-1}]\mathbf\Theta_{z_1}+
\tilde{\mathbf{A}}_d[\mathbf{X}_t,\mathbf{h}_{t-1}]\mathbf\Theta_{z_2}+\mathbf{b}_z) \\
\mathbf{r}_t \,&=\sigma (\tilde{\mathbf{A}}_s[\mathbf{X}_t,\mathbf{h}_{t-1}]\mathbf\Theta_{r_1}+
\tilde{\mathbf{A}}_d[\mathbf{X}_t,\mathbf{h}_{t-1}]\mathbf\Theta_{r_2}+\mathbf{b}_r) \\
\hat{\mathbf{h}}_t&=tanh(\tilde{\mathbf{A}}_s[\mathbf{X}_t,\mathbf{r}_t\odot\mathbf{h}_{t-1}]\mathbf\Theta_{\hat{\mathbf{h}}_1}  \\
&+\tilde{\mathbf{A}}_d[\mathbf{X}_t,\mathbf{r}_t\odot\mathbf{h}_{t-1}]\mathbf\Theta_{\hat{\mathbf{h}}_2}+ \mathbf{b}_{\hat{\mathbf{h}}}) \\
\mathbf{h}_t&=\mathbf{z}_t \odot \mathbf{h}_{t-1}+(1-\mathbf{z}_t) \odot\hat{\mathbf{h}}_t
\end{aligned}
\end{equation}
where $[,]$ denotes the concatenation operation, 
$\odot$ denotes element-wise multiplication,
$\mathbf{X}_t$ and $\mathbf{h}_t$ are input and output at time step $t$, $\mathbf{z}_t$ and $\mathbf{r}_t$ are reset gate and update gate, respectively.
$\mathbf\Theta_{\mathbf{z}_1}$, $\mathbf\Theta_{\mathbf{z}_2}$, $\mathbf\Theta_{\mathbf{r}_1}$,
$\mathbf\Theta_{\mathbf{r}_2}$, $\mathbf\Theta_{\hat{\mathbf{h}}_1}$, $\mathbf\Theta_{\hat{\mathbf{h}}_2}$, $\mathbf{b}_r$, $\mathbf{b}_r$, $\mathbf{b}_{\hat{\mathbf{h}}}$ are learnable parameters generated based on NAPL.

N-GCN have a multi-layer structure containing $L_n$ layers of SDGRU.
We take all hidden features of the last state $\mathbf{Z}_n \in \mathbf{R}^{L_n \times N\times F}$ as the final numerical features and feed $\mathbf{Z}_n$ into the subsequent fusion module.

\subsection{Vision-LSTM Network}
As shown in Figure~\ref{F1} (c), we introduce the V-LSTM to effectively capture spatio-temporal correlation from vision data. V-LSTM replaces the original MLP layer in LSTM with the proposed Multi-Scale joint Channel and Spatial Module (MSCSM), which is shown in Figure~\ref{F1} (d).

Taking $\mathbf{F}_{in}\in\mathbf{R}^{ h\times w\times c}$ as input, MSCSM first uses the Feature Pyramid Module (FPM), which consists of 1 $\times$ 1 convolution and three dilated convolution with different dilated rates.
FPM aims to capture multi-scale information $\mathbf{F}_{m} \in \mathbf{R}^{h\times w\times c}$ for recognizing different scale meteorological events.
Following that, MSCSM utilizes attention mechanisms along both channel and spatial dimensions for adaptive feature refinement.
For channel attention, we use average-pooling and max-pooling operations to generate $\mathbf{F}^c_{avg}$ and $\mathbf{F}^c_{max}$ to aggregate the spatial information of the feature graph, respectively.
Then a shared MLP is used to generate channel attention map $\mathbf{M}_c \in \mathbf{R}^{1\times1\times c}$.
For spatial attention, we use the same pooling operations across channels to generate $\mathbf{F}^s_{avg}$ and $\mathbf{F}^s_{max}$. 
The combination of the two is then fed into a standard convolutional layer to generate spatial attention map $\mathbf{M}_s \in \mathbf{R}^{ h\times w\times1}$.
The final feature map $\mathbf{M}_{a}$ is obtained by element-wise multiplication of $\mathbf{M}_c$ and $\mathbf{M}_s$, followed by a sigmoid activation. 
The final output of MSCSM $\mathbf{F}_{out}$ is computed as follow:
\begin{equation}
\resizebox{.91\linewidth}{!}{$
\begin{aligned}
\mathbf{F}_{m}&=FPM(\mathbf{F}_{in}) \\
\mathbf{M}_c&=MLP(AvgPool(\mathbf{F}_{m}))+MLP(MaxPool(\mathbf{F}_{m})) \\
&=MLP(\mathbf{F}^c_{avg})+MLP(\mathbf{F}^c_{max}) \\
\mathbf{M}_s&=Conv2D([AvgPool(\mathbf{F}_{m}),MaxPool(\mathbf{F}_{m})]) \\ 
&=Conv2D([\mathbf{F}^s_{avg},\mathbf{F}^s_{max}]) \\
\mathbf{M}_{a}&=\sigma(\mathbf{M}_c\odot\mathbf{M}_s)\\
\mathbf{F}_{out}&=\mathbf{F}_{m}\odot\mathbf{M}_{a} 
\end{aligned}
$}
\end{equation}
The V-LSTM determines the future state of a certain cell by the inputs and past states of its local neighbors.
The calculation process in V-LSTM is as follows:
\begin{equation}
\begin{aligned}
\mathbf{F}_{mscs}&=MSCSM([\mathbf{I}_t,\mathbf{H}_{t-1}]) \\
\mathbf{i}_t \enspace &=\sigma(\mathbf{W}_i*\mathbf{F}_{mscs}+\mathbf{b}_i) \\
\mathbf{f}_t \enspace &=\sigma(\mathbf{W}_f*\mathbf{F}_{mscs}+\mathbf{b}_f) \\
\mathbf{c}_t \; &=\mathbf{f}_t \odot \mathbf{c}_{t-1} + \mathbf{i}_t \odot tanh(\mathbf{W}_c*\mathbf{F}_{mscs} +\mathbf{b}_c) \\
\mathbf{o}_t \:&=\sigma(\mathbf{W}_o*\mathbf{F}_{mscs} +\mathbf{b}_o) \\
\mathbf{H}_t&=\mathbf{o}_t \odot tanh(\mathbf{c}_t) \\
\end{aligned}
\end{equation}
where $*$ denotes the convolution operator.
The Vision-LSTM Network is constructed by stacking $L_v$ layers of V-LSTM with down-sampling operations between layers.
We take hidden features from the final state of the last layer $\mathbf{Z}_v \in \mathbf{R}^{H^{'}\times W^{'} \times F}$ as the final vision features and feed $\mathbf{Z}_v$ into the subsequent fusion module.
$H^{'}$ and $W^{'}$ are height and width of the output feature.

\subsection{Vision-Numerical Fusion}
Double Query Attention Module (DQAM) aims to conduct multi-modal feature interaction using numerical features $\mathbf{Z}_n$ extracted by N-GCN and vision features $\mathbf{Z}_v$ extracted by V-LSTM network as inputs.
First, $\mathbf{Z}_n$ and $\mathbf{Z}_v$ are  formulated as $\mathbf{Z}^{'}_n \in \mathbf{R}^{L_{n}N\times F}$ and $\mathbf{Z}^{'}_v \in \mathbf{R}^{H^{'}W^{'} \times C^{'}}$  through matrix manipulation. 
DQAM is illustrated as follows:
\begin{equation}
\resizebox{.89\linewidth}{!}{$
\begin{aligned}
\mathbf{Att}&=softmax(\frac{(\mathbf{W}_Q[\mathbf{Z}^{'}_n,\mathbf{Z}_L])(\mathbf{W}_K\mathbf{Z}^{'}_v)^T}{\sqrt{F}})\mathbf{W}_V\mathbf{Z}^{'}_v \\
\mathbf{O}&=LayerNorm(\mathbf{Z}^{'}_n+\mathbf{Att})
\end{aligned}
$}
\end{equation}
where $\mathbf{Z}_L \in \mathbf{R}^{L_{n}N\times F}$ is the additional learnable query.
$\mathbf{W}_Q$, $\mathbf{W}_K$ and $\mathbf{W}_V$ are linear parameters, transforming the dimensions into $F$.
The attention output $\mathbf{O} \in \mathbf{R}^{L_{n}N\times F} $ is reshaped to $\mathbf{R}^{L_{n}\times N\times F}$, split into $L_n$ parts, and fed into the GCN-based decoder to generate hourly predictions of specified meteorological factors.

\begin{table*}[htb]
\scriptsize
\renewcommand{\thetable}{1}
\begin{center}
\tabcolsep= 0.13cm
\renewcommand\arraystretch{1.2}
\begin{tabular}{l|c c|c c|c c}
\toprule
\textbf{Factor} & \multicolumn{2}{c|}{Temperature}& \multicolumn{2}{c|}{Relative Humidity} & \multicolumn{2}{c}{Visibility} \\ \midrule

\textbf{Model} & MAE & RMSE & MAE & RMSE & MAE & RMSE  \\ \midrule
ASTGCN & 2.2509 / 1.4003 / 1.5037 & 3.0200 / 1.9695 / 2.0270 & \enspace 9.8476 / 7.5614 / 7.3076 & 13.3175 / 10.3637 / 10.0980 & 4.4816 / 4.2856 / 4.6761 & 6.8576 / 6.5775 / 6.3955 \\
MSTGCN & 2.8056 / 1.8616 / 1.8769 & 3.7427 / 2.5534 / 2.5311 & 11.8937 / 8.8681 / 9.5984 & 15.6763 / 11.9260 / 12.8901 & 5.6741 / 4.9069 / 5.1959 & 7.7556 / 6.9803 / 6.8882  \\
TGCN & 1.7534 / 1.3091 / 1.2362 & 2.3970 / 1.8490 / 1.7145 & \enspace 8.9419 / 7.0038 / 6.9915 & 12.1712 / \enspace 9.6341 / \enspace 9.6930 & 4.9798 / 4.4764 / 4.5413 & 7.5513 / 6.8084 / 6.3250 \\ 
AGCRN & 1.6528 / 1.1097 / 1.2405 & 2.2705 / 1.5042 / 1.7198 & \enspace 7.9552 / 5.8377 / 5.9908 & 11.1039 / \enspace 8.3471 / \enspace 8.6770 & 4.2666 / 3.8276 / 3.8537 & 6.7296 / 6.0791 / 5.5422  \\ 
CLCSTN & 2.2357 / 1.4159 / 1.5339 & 3.0121 / 1.9895 / 2.0878 & 10.4855 / 7.5468 / 7.9679 & 14.2840 / 10.3815 / 10.9861 & 5.0497 / 4.4903 / 4.7450 & 7.5226 / 6.8436 / 6.4239  \\ 
CLCRN & 1.6073 / \underline{0.9455} / 1.0623 & 2.2454 / \underline{1.4218} / 1.5248 & \enspace 7.8973 / 5.6311 / 5.9542 & 11.3483 / \enspace 8.1954 / \enspace 8.5480 & 4.1264 / \underline{3.6441} / 3.7653 & 6.7191 / \underline{5.9336} / 5.5382 \\ 
MFMGCN & \underline{1.4711} / 0.9715 / \underline{0.9958} & \underline{2.0520} / 1.4330 / \underline{1.4386} & \enspace \underline{7.4334} / \underline{5.4985} / \underline{5.7474} & \underline{10.5738} / \enspace \underline{7.9297} / \enspace \underline{8.3128} & \underline{4.0242} / 3.6762 / \underline{3.7406} & \underline{6.6041} / 5.9719 / \underline{5.4779}  \\ \midrule
VN-Net & \textbf{1.3889} / \textbf{0.9126} / \textbf{0.9805} & \textbf{1.9066} / \textbf{1.3554} / \textbf{1.4335} & \enspace \textbf{7.0364} / \textbf{5.3836} / \textbf{5.5852} & \textbf{10.0453} / \enspace
\textbf{7.7910} / \enspace \textbf{8.0568} & \textbf{3.9015} / \textbf{3.6366} / \textbf{3.7063} & \textbf{6.3487} / \textbf{5.9202} / \textbf{5.4062} \\   
\bottomrule
\end{tabular}
\caption{Overall comparison between VN-Net and baseline methods. 
MAE and RMSE are presented by regions as follows: Northeast / Southwest / Southeast. Results with \underline{underlines} are the best performance achieved by baselines, and results with \textbf{bold} are the overall best.} 
\label{t1}
\end{center}
\end{table*}
\begin{figure}
\centerline{\includegraphics[scale=0.85]{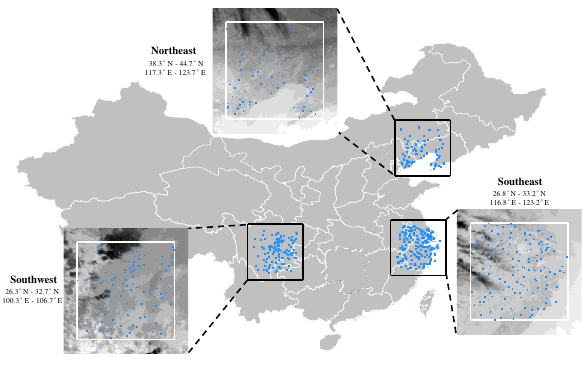}}
\caption{Three regions of the datasets. Northeast (38.3°N-44.7°N, 117.3°E-123.7°E), Southwest (26.3°N-32.7°N, 100.3°E-106.7°E), Southeast (26.8°N-33.2°N, 116.8°E-123.2°E) with 60, 96, and 139 ground weather stations, respectively.}
\label{f2}
\end{figure}
\subsection{Meteorological Factors Forecasting}
The GCN-based decoder also has $L_n$ layers and shares a similar structure with N-GCN, except that it does not incorporate time embeddings and dynamic dependency construction.

The decoder is initialized by the output of DQAM and generates predictions given previous ground truth signals during training, while the ground truth signals are replace by predictions generated by the decoder itself during testing.
To eliminate the possible performance degradation caused by differences in data distribution between the training and testing, we integrate scheduled sampling~\cite{bengio2015scheduled} into the model.
During training, the decoder makes predictions based on previous ground truth with probability $p_i$ or the decoder output with probability $1-p_i$ at i-th mini-batch.
We use an inverse sigmoid decay schedule for $p_i$ in the scheduled sampling approach:
\begin{equation}
p_i=\frac{k}{k+e^{\frac{i}{k}}}
\end{equation}
where $k\geq1$ depends on the expected speed of convergence.

\nocite{bengio2015scheduled}

\section{Experiments}
\subsection{Datasets}
Weather2K~\cite{zhu2023weather2k} is a very recent benchmark dataset for spatio-temporal meteorological forecasting based on hourly observation data from 1,866 ground weather stations, with 3 static variables representing geographic information and 20 interacting meteorological factors.
Himawari-8 is a geostationary meteorological satellite which carries state-of-the-art payload called Advanced Himawari Imager (AHI) with 16 observation bands~\cite{bessho2016introduction}.

As shown in Figure~\ref{f2}, Weather2K and Himawari-8 provide numerical data and vision image data, respectively.
Considering the influence of different geographical locations and density of stations, we extract the region of northeast (NE, 39.0°N - 44.0°N, 118.0°E - 123.0°E), southwest (SW, 27.0°N - 32.0°N, 101.0°E - 106.0°E), and southeast (SE, 27.5°N - 32.5°N, 117.5°E - 122.5°E) from Weather2K, which contain 60, 96, and 139 ground weather stations, respectively.
Due to the requirement for full time coverage observation, we choose Himawari-8 infrared bands from 11 to 16.
The spatial coverage of satellite images in each region is 6.4° $\times$ 6.4°, providing a wider vision information.
See more details of dataset in Supplementary Material A.
\nocite{zhu2023weather2k}
\nocite{bessho2016introduction}

\subsection{Implementation Details}
\subsubsection{Baselines}
We selected seven state-of-the-art ST-GNN models as baselines: ASTGCN ~\cite{guo2019attention}, MSTGCN~\cite{guo2019attention}, TGCN~\cite{zhao2019t}, AGCRN~\cite{bai2020adaptive}, CLCSTN~\cite{lin2022conditional}, CLCRN~\cite{lin2022conditional}, MFMGCN~\cite{zhu2023weather2k}.
\nocite{guo2019attention}
\nocite{zhao2019t}
\nocite{bai2020adaptive}
\nocite{lin2022conditional}
\nocite{zhu2023weather2k}

\subsubsection{Hyper-parameters}
We set $T_h=T_p=12$, which means both the input time length and forecasting time length are 12 steps. 
The $H$ and $W$ of vision data are both 160, down-sampling from the original size.
We set $L_n=2$ and $L_v=3$ to construct encoders.
The dimension of hidden feature $F=32$ and embedding $d_e=16$.
We set $k=1000$ in scheduled sampling.

All the models are trained with the loss function of MAE and optimized by Adam optimizer for 100 epochs.
Early-stopping epoch is set to 30. The batch size for training is set to 16.
The initial learning rate is set to $1e^{-2}$, and it decays with the ratio $5e^{-2}$ per 10 epochs in the first 50 epochs.

\subsubsection{Metrics}
We compare VN-Net with other methods by deploying mean absolute error (MAE) and root mean square error (RMSE) to measure the performance of forecasting models.
In addition, we introduce the metrics of Top-5 Meteorological Factors Contribution (MFC) and Static Information Contribution (SIC) in interpretation analysis.

\begin{table*}
\tabcolsep= 0.25cm
\renewcommand{\thetable}{2}
    \centering
    \begin{tabular}{l c c c c}
    \toprule
        \textbf{Numerical Branch} & \textbf{Vision Branch} & \textbf{Attention Query} & \textbf{MAE} & \textbf{RMSE} \\ 
        \midrule
        \qquad N-GCN* & - & - & 1.4909 / 7.4218 / 4.0397 & 2.0725 / 10.5586 / 6.6245 \\ 
        \qquad N-GCN & - & - & 1.4595 / 7.2923 / 4.0313 & 2.0399 /  10.3170 / 6.6044 \\ 
        \qquad N-GCN & ConvLSTM & Single & 1.4258 /  7.1430 / 4.0177 & 1.9706 / 10.2661 / 6.6040 \\ 
        \qquad  N-GCN & V-LSTM & Single & 1.3996 / 7.0569 / 3.9542 & 1.9336 / 10.1117 / 6.5257 \\ 
        \qquad N-GCN & V-LSTM & Double & 1.3889 / 7.0364 / 3.9015 & 1.9066 / 10.0453 / 6.3487 \\ 
        \bottomrule
    \end{tabular}
    \caption{The results of ablation study of the NE region. 
    MAE and RMSE are represented by meteorological factors as follows: temperature / relative humidity / visibility. N-GCN* represents N-GCN without time embedding.} 
    \label{t2}
\end{table*}
\begin{figure*}[!ht] 
\centerline{\includegraphics[scale=0.7]{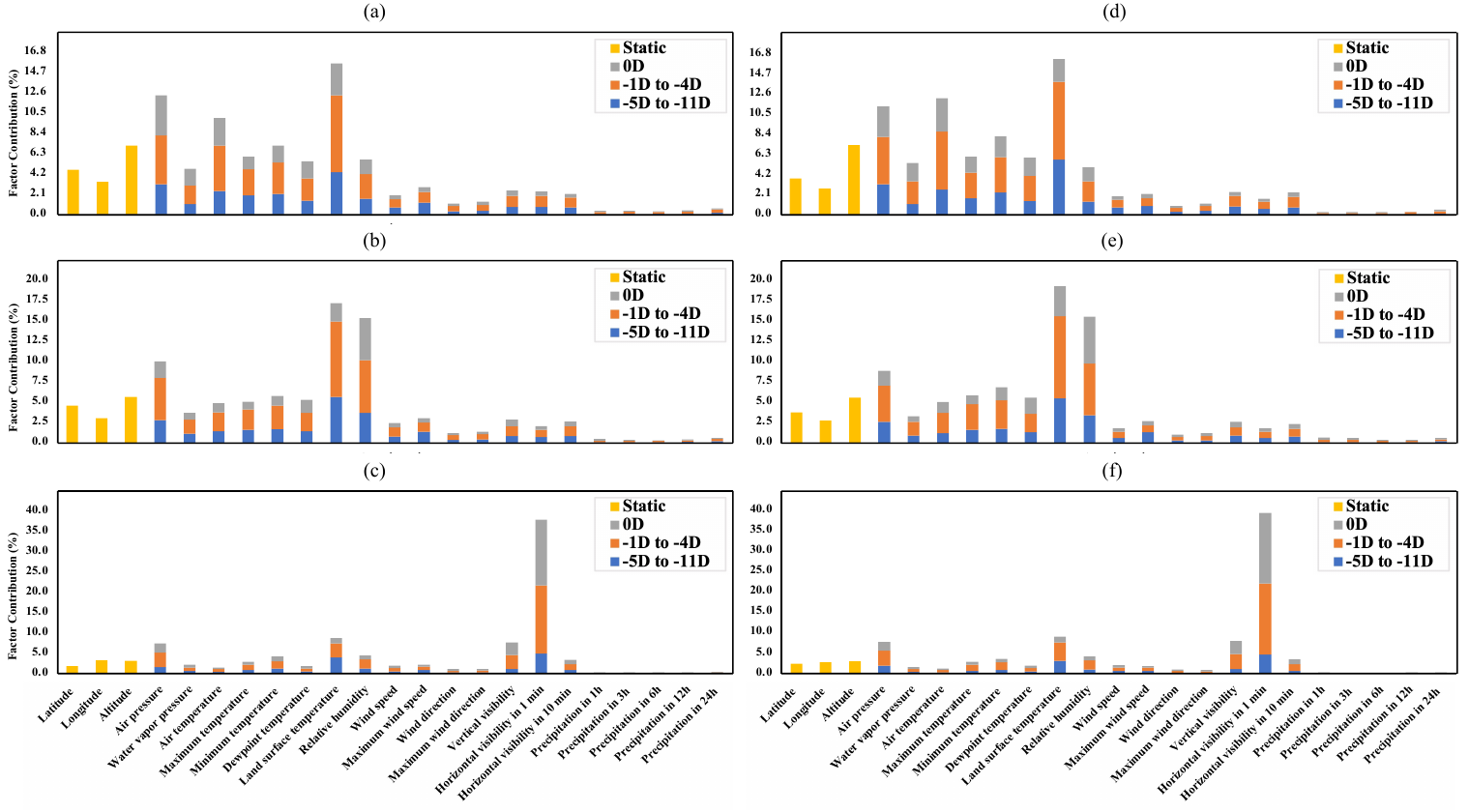}}
\caption{Factor contribution results of the SW region. (a) Temperature, uni-modal, (b) Relative Humidity, uni-modal. (c) Visibility, uni-modal. (d) Temperature, multi-modal, (e) Relative Humidity, multi-modal. (f) Visibility, multi-modal. Factor names follow Weather2K notation. The legend label -$i$D and Static define the data contribution of the $i$ days before forecast initialization and 3 geolocation information constants, respectively.}
\label{f3}
\end{figure*}

\subsection{Results}
\subsubsection{Overall Comparison}
The results of overall comparison are shown in Table~\ref{t1}.
Notably, VN-Net outperforms state-of-the-art by a significant margin in both different regions and meteorological factors, which
demonstrates the generality of our framework and effectiveness of combining the vision data with the numerical data for sparse spatio-temporal meteorological forecasting.

In the perspective of region-based experimental analysis, VN-Net brings different relative improvements to the existing best results of three factors.
Specifically, VN-Net achieves an average gain of 4.66\% and 5.32\%, 1.93\% and 2.92\%, 1.76\% and 1.58\% on MAE and RMSE in the region of NE, SW, and SE, respectively.
We find that the performance gain is inversely proportional to the number of ground weather stations, which indicates the utilization of vision data may be related to the density of stations.

\subsubsection{Ablation Study}
In order to better verify the effectiveness of several key modules in our VN-Net, we conduct ablation experiments and present the results of Northeast region in Table~\ref{t2}.
The experiment results show that: (1) In the numerical data branch, we find that the design of time embedding brings great gains to N-GCN.
(2) In the vision data branch, V-LSTM shows excellent ability to extract visual information from multi-channel time series satellite images.
We also provide an experimental setup using ConvLSTM~\cite{shi2015convolutional}, which eliminates the bias that performance gains are simply due to the use of multi-modal data.
(3) The addition of learnable query in DQAM can better integrate vision feature under the guidance of numerical feature.
See ablation study of other regions in Supplementary Material B.
\nocite{shi2015convolutional}

\begin{table}
\renewcommand{\thetable}{3}
\small
    \centering
    \begin{tabular}{l c c c}
        \toprule
        \textbf{Factor} & \textbf{Region} & \textbf{Top-5 MFC (\%)} & \textbf{SIC (\%)} \\ \midrule
    
        \multirow{3}{*}{Temperature} & NE & +3.00 & -1.35  \\ 
        & SW & +3.03 & -1.34  \\ 
        & SE & +4.21 & -1.37  \\ \midrule
        \multirow{3}{*}{Relative Humidity} & NE & +0.48 & -1.26  \\ 
        & SW & +2.68 & -1.32  \\ 
        & SE & +2.02 & -1.44  \\ \midrule
        \multirow{3}{*}{Visibility} & NE & +3.83 & -1.00  \\ 
        & SW & +2.15 & -0.19  \\ 
        & SE & +1.92 & -0.12 \\ 
        \bottomrule
    \end{tabular}
    \caption{The change of Top-5 MFC and SIC after the introduction of vision data. + and - represent increase and decrease, respectively.} 
    \label{t3}
\end{table}

\section{Interpretation}
It is important to understand what the physics-free neural network is learning, especially in the case of multivariate variables as inputs.
In this section, we aim to gain new insights into the interactions between meteorological factors before and after the introduction of vision data, and ensure that the model conforms to prior knowledge of weather physics.

We assess the contribution of multivariate meteorological factors using integrated gradients~\cite{sundararajan2017axiomatic}.
Specifically, we use the mean value of each meteorological factor and the minimum value of each channel as baseline input $\mathbf{X}^{b}$ and $\mathbf{I}^{b}$, which result in neutral forecasting.
The contribution of each input time $\mathbf{IG}_t$ is computed as the partial derivative of the forecasting with respect to $\mathbf{X}_t$, integrated along a linear path from the baseline to the actual input:
\begin{equation}
\mathbf{IG}_t=\frac{1}{T_p \cdot N^2} \left\| \int_{\alpha=0}^1 \frac{\partial \theta\left(\mathbf{X}_\alpha,\mathbf{I}_\alpha\right)}{\partial \mathbf{X}_t} d \alpha
\left(\mathbf{X}_{t}-\mathbf{X}_{t}^b\right)\right\|_1
\end{equation}
where $\mathbf{X}_\alpha=\mathbf{X}^{b}+\alpha \times\left(\mathbf{X}^{h}-\mathbf{X}^{b}\right)$,  
$\mathbf{I}_\alpha=\mathbf{I}^{b}+\alpha \times\left(\mathbf{I}^{h}-\mathbf{I}^{b}\right)$.
The contribution of all input data can be expressed as $\mathbf{IG}^h=[\mathbf{IG}_{t-{T_h}+1}$, $\cdots,\mathbf{IG}_t] \in \mathbf{R}^{ T_h\times D}$.
To eliminate the bias of absolute attribution values of different forecasting tasks, we convert factor contributions to percentages as follows:
\begin{equation}
IG_{\left(t: t+i, f*\right)}=\sum_t^{t+i} \mathbf{IG}^h\left[t ; f*\right] / {\operatorname{SUM}\left(\mathbf{IG}^h\right)} \times 100\%
\end{equation}
where $IG_{\left(t: t+i, f*\right)}$ represents the contribution rate of meteorological factor $f^*$ from time $t$ to $t+i$.
We present the results of the SW region using uni-modal and multi-modal data in Figure~\ref{f3}.
The contributions of the most recent data, previous four hours data and the first seven hours data are shown in different colors to quantify the relevance of past history as a predictor of future states.
See results of other regions in Supplementary Material C.

To compare the difference between uni-modal and multi-modal, we specifically introduce the metrics of Top-5 MFC and SIC, which represent the contribution ratio of the top 5 effective meteorological factors and static information (latitude, longitude, and altitude), respectively.
Table~\ref{t3} shows the changes in these two metrics after the introduction of vision data.
By analyzing the results of Figure~\ref{f3} and Table~\ref{t3}, we can conclude that:
(1) The meteorological factor itself is the most or equally important features in all cases.
(2) The impact of the most recent data is more significant than any day in the input data.
(3) For a specific task, the top 5 effective meteorological factors remain unchanged regardless of uni-modal or multi-modal input.
It is worth noting that Top-5 MFC maintains a positive growth in multi-modal cases, indicating that VN-Net improves the utilization of key factors.
(4) The decrease of SIC in multi-modal cases indicates that satellite vision data complements the geolocation information.

\nocite{sundararajan2017axiomatic}


\section{Conclusion}
In this paper, we present a novel multi-modal graph model named VN-Net to solve sparse spatio-temporal meteorological forecasting.
Specifically, VN-Net utilizes two branches: N-GCN processes numerical data from ground weather stations, while another, V-LSTM processes visual data from meteorological satellites.
A multi-modal fusion module called DQAM is developed to integrate vision features under the guidance of numerical features, and then the fusion features generate final predictions through the GCN-based decoder.
Extensive experiments with various regions and meteorological factors on new released Weather2K dataset demonstrate the generality of VN-Net for enhancing forecasting performance.
Furthermore, we conduct interpretation analysis to gain new insights into the changes of the meteorological factor contributions of before and after the introduction of vision data.
We hope that the proposal of VN-Net can motivate more researches on the potential of using multi-modal data in sparse meteorological forecasting.

\bibliographystyle{named}
\bibliography{ijcai24}

\newpage
\renewcommand\thesection{\Alph{section}}
\onecolumn
\textbf{\centerline{\fontsize{14}{0}\selectfont Supplementary Materials to}}
\textbf{\centerline{\fontsize{14}{0}\selectfont VN-Net: Vision-Numerical Fusion Graph Convolutional Network}}
\textbf{\centerline{\fontsize{14}{0}\selectfont for Sparse Spatio-Temporal Meteorological Forecasting}}
\appendix

\section{Dataset}
\subsection{Weather2K}
The Weather2K dataset provides observational variables of 1,866 ground weather stations from January 2017 to August 2021 with a temporal resolution of 1 hour. 
For each station, 20 meteorological variables and 3 constants for position information are recorded, as shown in Table~\ref{t4}. 
We use all multivariate variables as the input of numerical data and select three meteorological factors as the focus of our research, which are air temperature, relative humidity, and horizontal visibility in 1 min.

To verify the model’s generalization performance at different geographical locations and density of stations, we extract the region of northeast (39.0°N - 44.0°N, 118.0°E - 123.0°E), southwest (27.0°N - 32.0°N, 101.0°E - 106.0°E), and southeast (27.5°N - 32.5°N, 117.5°E - 122.5°E) from Weather2K, which contain 60, 96, and 139 ground weather stations, respectively.

\begin{table*}[!th]
\scriptsize
\renewcommand{\thetable}{4}

\begin{center}
\begin{tabular}{l|c|l}
\toprule
\textbf{Long Name} &\textbf{Unit} & \textbf{Physical Description} \\ \midrule
Latitude & (°) & The latitude of the ground observation station \\ \hline
Longitude & (°) & The longitude of the ground observation station \\ \hline
Altitude & (m) & The altitude of the air pressure sensor \\ \bottomrule
Air pressure  & hpa & Instantaneous atmospheric pressure at 2 meters above the ground \\ \hline
Water vapor pressure & hpa & Instantaneous partial pressure of water vapor in the air \\ \hline
Air temperature & (°C) &  Instantaneous temperature of the air at 2.5 meters above the ground where sheltered from direct solar radiation \\ \hline
Maximum / Minimum temperature & (°C) & Maximum / Minimum air temperature in the last one hour \\ \hline
Dewpoint temperature & (°C) & Instantaneous temperature at which the water vapor saturates and begins to condense\\ \hline
Land surface temperature & (°C) & Instantaneous temperature of bare soil at the ground surface \\ \hline
Relative humidity & (\%) & Instantaneous humidity relative to saturation at 2.5 meters above the ground\\ \hline
Wind speed & ($\mathrm{ms^{-1}}$) & The average speed of the wind at 10 meters above the ground in a 10-minute period  \\ \hline
Maximum wind speed  & ($\mathrm{ms^{-1}}$) & Maximum wind speed in the last one hour \\ \hline
Wind direction & (°) & The direction of the wind speed. (Wind direction is 0 if wind speed is less than or equal to 0.2)\\ \hline
Maximum wind direction & (°) & Maximum wind speed's direction in the last one hour \\ \hline
Vertical visibility & (m) & Instantaneous vertical visibility \\ \hline
Horizontal visibility in 1 min / 10 min & (m) & 1 / 10 minute(s) mean horizontal visibility at 2.8 meters above the ground \\ \hline
Precipitation in 1h / 3h / 6h / 12h / 24h  & (mm) & Cumulative precipitation in the last 1 / 3 / 6 / 12 / 24 hour(s)\\ \bottomrule
\end{tabular}
\caption{Multivariate variables of the Weather2K dataset.} 
\label{t4}
\end{center}
\end{table*}

\subsection{Himawari-8}
Himawari-8 is a geostationary satellite operated by the Japan Meteorological Agency (JMA), which has a new payload called the Advanced Himawari Imager (AHI).
AHI has 16 spectral bands with a spatial resolution of up to 2 km and a temporal resolution of 10 minutes.
For full time coverage observation, we choose Himawari-8 infrared bands from 11 to 16. 

We download Himawari-8 gridded data through the FTP server\footnote{\url{ftp://hmwr829gr.cr.chiba-u.ac.jp}} provided by Center for Environmental Remote Sensing (CEReS) of Chiba University, Japan.
We choose Version 02 (V20190123) for high geo-correction accuracy.
Notably, CEReS gridded data has different bands naming rule with JMA official Himawari 8 bands names, as shown in Table~\ref{t5}.

\begin{table}[!th]
\renewcommand{\thetable}{5}
\begin{center}
\tabcolsep= 0.32cm
\begin{tabular}{l|c|c|c}
\toprule
\textbf{JMA AHI Band} & \textbf{CEReS Gridded data} & \textbf{Pixel * Line} & \textbf{Spatial Resolution} \\ \midrule
Band 11 (8.6 \si{\micro\metre}) & TIR 09 & \multirow{6}{*}{6000 * 6000} & \multirow{6}{*}{0.02 Degree(approx. 2km)}\\
Band 12 (9.6 \si{\micro\metre}) & TIR 10 &  & \\
Band 13 (10.4 \si{\micro\metre}) & TIR 01 &  & \\
Band 14 (11.2 \si{\micro\metre}) & TIR 02 &  & \\
Band 15 (12.4 \si{\micro\metre}) & TIR 03 &  & \\
Band 16 (13.3 \si{\micro\metre}) & TIR 04 &  & \\
\bottomrule
\end{tabular}
\caption{Band relation between JMA AHI and CEReS gridded data. TIR represents thermal infrared.} 
\label{t5}
\end{center}
\end{table}

Following the visualization process of CEReS, we get vision data of northeast (38.3°N - 44.7°N, 117.3°E - 123.7°E), southwest (26.3°N - 32.7°N, 100.3°E - 106.7°E), and southeast (26.8°N - 33.2°N, 116.8°E - 123.2°E) with a wider region than numerical data.
Specially, we first download the file compressed in bz2 archive format.
After uncompressing, the data file type is unsigned short integer, big endian binary.
Then we use a byte swap option to get little endian data and the value of the raw digital data is distributed in the range of 1 to 4096.
According to the calibration table produced by JMA, we convert the count value into the physical variable of brightness temperature (Tbb).
In addition, it is optional to utilize Generic Mapping Tools to generate grayscale image for visualization.
Finally, we crop the specified region and store in Numpy format file.

\subsection{Dataset Split}
Weather2K and Himawari-8 datasets are divided into three parts in chronological order: data from January 1, 2017, to August 31, 2019, is used for training; data from September 1, 2019, to August 31, 2020, is used for validation; and data from September 1, 2020, to August 31, 2021, is used for testing. 
We conduct interpretation analysis on test set.

\section{Ablation Study}
\subsection{SW Region}
Table~\ref{t6} shows the ablation study of the SW region. 
Here we have corrected a data error in Table 1 of the main content: the RMSE of relative humidity of SW region of VN-Net is 7.7910, instead of 7.9120. 
(The original result was 7.79118, but due to a mistake recorded, it was changed to 7.9118 and rounded to 7.9120.)
The relevant analysis in the main content is based on the correct 7.7910, so it is correct. Sincerely apologize.
\begin{table*}[!th]
\tabcolsep= 0.25cm
\renewcommand{\thetable}{6}
    \centering
    \begin{tabular}{l c c c c}
    \toprule
        \textbf{Numerical Branch} & \textbf{Vision Branch} & \textbf{Attention Query} & \textbf{MAE} & \textbf{RMSE} \\ 
        \midrule
        \qquad N-GCN* & - & - & 0.9460 / 5.5875	/ 3.6801 & 1.4212 / 8.0194 / 5.9868 \\ 
        \qquad N-GCN & - & - & 0.9389 / 5.4847 / 3.6587 & 1.4077 / 7.9533 / 5.9736 \\ 
        \qquad N-GCN & ConvLSTM & Single & 0.9355 / 5.4800 / 3.6685 & 1.3889 / 7.9441 / 5.9778 \\ 
        \qquad  N-GCN & V-LSTM & Single & 0.9257 / 5.4222 / 3.6472 & 1.3739 / 7.8087 / 5.9554  \\ 
        \qquad N-GCN & V-LSTM & Double & 0.9126 / 5.3836 / 3.6366 & 1.3554 / 7.7910 / 5.9202 \\ 
        \bottomrule
    \end{tabular}
    \caption{The results of ablation study of the SW region. 
    MAE and RMSE are represented by meteorological factors as follows: temperature / relative humidity / visibility. N-GCN* represents N-GCN without time embedding.} 
    \label{t6}
\end{table*}

\subsection{SE Region}
Table~\ref{t7} shows the ablation study of the SE region.
\begin{table*}[!th]
\tabcolsep= 0.25cm
\renewcommand{\thetable}{7}
    \centering
    \begin{tabular}{l c c c c}
    \toprule
        \textbf{Numerical Branch} & \textbf{Vision Branch} & \textbf{Attention Query} & \textbf{MAE} & \textbf{RMSE} \\ 
        \midrule
        \qquad N-GCN* & - & - & 1.0141 / 5.7508 / 3.7444 & 1.4578 / 8.3162 / 5.5162 \\ 
        \qquad N-GCN & - & - & 1.0080 / 5.7391 / 3.7402 & 1.4544 / 8.3063 / 5.4765 \\ 
        \qquad N-GCN & ConvLSTM & Single & 1.0041 / 5.7385 / 3.7350 & 1.4573 / 8.2675 / 5.4446 \\ 
        \qquad  N-GCN & V-LSTM & Single & 0.9846 / 5.6389 / 3.7103 & 1.4339 / 8.1493 / 5.4426 \\ 
        \qquad N-GCN & V-LSTM & Double & 0.9805 / 5.5852 / 3.7063 & 1.4335 / 8.0568 / 5.4062 \\ 
        \bottomrule
    \end{tabular}
    \caption{The results of ablation study of the SE region. 
    MAE and RMSE are represented by meteorological factors as follows: temperature / relative humidity / visibility. N-GCN* represents N-GCN without time embedding.} 
    \label{t7}
\end{table*}

\section{Interpretation}
\subsection{NE Region}
\begin{figure*}[!ht] 
\renewcommand{\thefigure}{4}
\centerline{\includegraphics[scale=0.7]{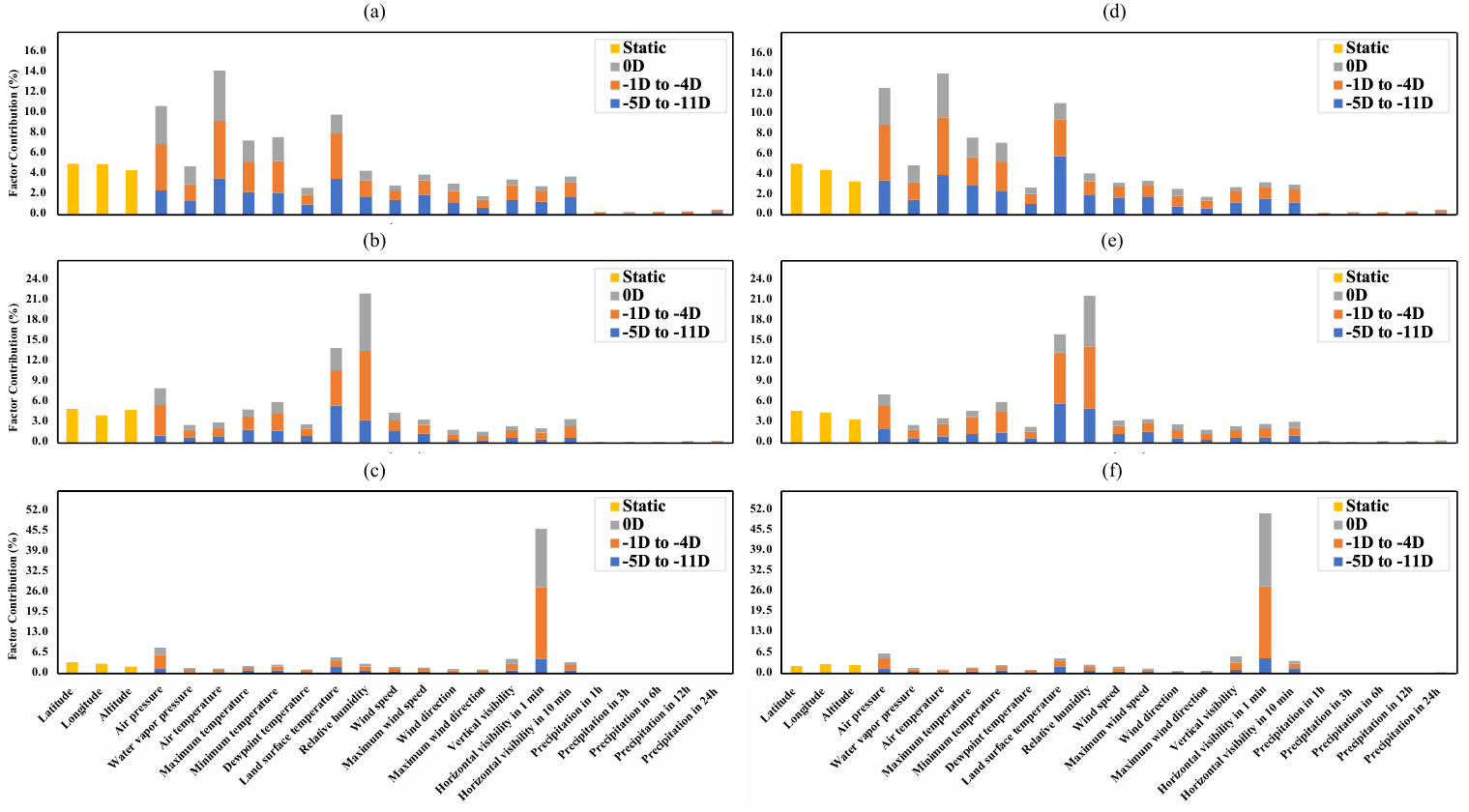}}
\caption{Factor contribution results of the NE region. (a) Temperature, uni-modal, (b) Relative Humidity, uni-modal. (c) Visibility, uni-modal. (d) Temperature, multi-modal, (e) Relative Humidity, multi-modal. (f) Visibility, multi-modal.}
\label{f4}
\end{figure*}

Figure~\ref{f4} shows the factor contribution results of the NE region using uni-modal and multi-modal data.


\subsection{SE Region}
\begin{figure*}[!ht] 
\renewcommand{\thefigure}{5}
\centerline{\includegraphics[scale=0.7]{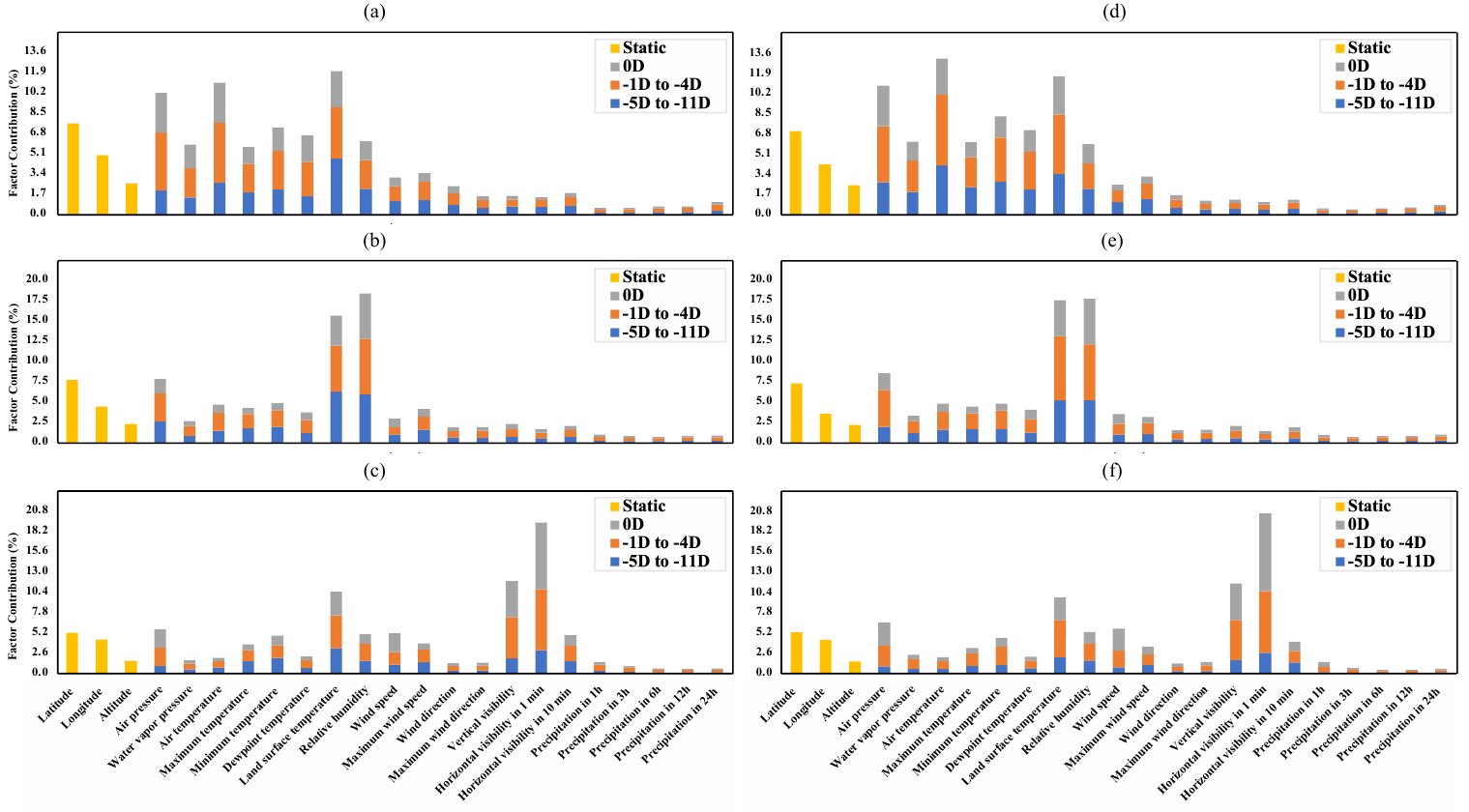}}
\caption{Factor contribution results of the SE region. (a) Temperature, uni-modal, (b) Relative Humidity, uni-modal. (c) Visibility, uni-modal. (d) Temperature, multi-modal, (e) Relative Humidity, multi-modal. (f) Visibility, multi-modal.}
\label{f5}
\end{figure*}

Figure \ref{f5} shows the factor contribution results of the SE region using uni-modal and multi-modal data.


\end{document}